\documentclass[]{fairmeta}

\usepackage[utf8]{inputenc}
\usepackage[T1]{fontenc}
\usepackage{amsmath}
\usepackage{amssymb}
\usepackage{tabularx}
\IfFileExists{enumitem.sty}{\usepackage{enumitem}}{}
\IfFileExists{comment.sty}{\usepackage{comment}}{\newenvironment{comment}{\par\iffalse}{\fi}}
\usepackage{pifont}
\usepackage{xspace}
\graphicspath{{../figures/}}

\newcommand{\ours}{\textsc{SWE-Together}\xspace}

\newcommand{\tabref}[1]{Table~\ref{#1}}
\newcommand{\figref}[1]{Fig.~\ref{#1}}

\newcommand{\cmark}{\textcolor{green!50!black}{\ding{51}}}   
\newcommand{\xmark}{\textcolor{red!65!black}{\ding{55}}}      
\newcommand{\pmark}{\textcolor{orange!85!black}{$\blacktriangle$}} 
\newcommand{\mmark}{\textcolor{blue!65!black}{$\blacklozenge$}}    

\title{\ours: Evaluating Coding Agents in Interactive User Sessions}

\author[*]{Yifan Wu}
\author{Zhuokai Zhao}
\author{Songlin Li}
\author{Ho Hin Lee}
\author{Jiacheng Zhu}
\author{Shirley Wu}
\author{Tianhe Yu}
\author{Serena Li}
\author{Lizhu Zhang}
\author{Xiangjun Fan}
\author[*]{Shengzhi Li}
\affiliation{Meta}
\contribution[*]{Project Co-Lead}

\abstract{Most coding-agent benchmarks are static: an agent receives a complete task description up front and is judged only by its final code. Real coding assistance is interactive, with users clarifying goals, adding constraints, and correcting mistakes over multiple turns. We introduce \textbf{SWE-Together}, a multi-turn benchmark reconstructed from real user--agent coding sessions. To make real interactions verifiable, we curate 109 repository-level tasks from 11{,}260 recorded sessions, selecting sessions with recoverable repository states, clear user goals, and observable outcomes. To replay these interactions across agents, we build a reactive LLM-based user simulator that preserves the original users' intents and provides feedback when the coding agent's progress requires it. To evaluate agents as collaborators, we measure both final repository correctness and the number of corrective feedback turns required during the interaction. Experiments with frontier coding agents show that stronger agents generally achieve higher final success rates while requiring fewer interventions, suggesting an improved user experience.

}

\date{\today}
\correspondence{
  Yifan Wu at \href{https://yifannnwu.com}{yifannnwu.com}
  and Shengzhi Li at \href{https://lishengzhi.com}{lishengzhi.com}
}
\metadata[Code]{\url{https://github.com/Togetherbench/SWE-Together}}
\metadata[Website]{\url{https://togetherbench.com}}

\begin{document}
\maketitle
\begingroup
\centering
\includegraphics[width=\linewidth]{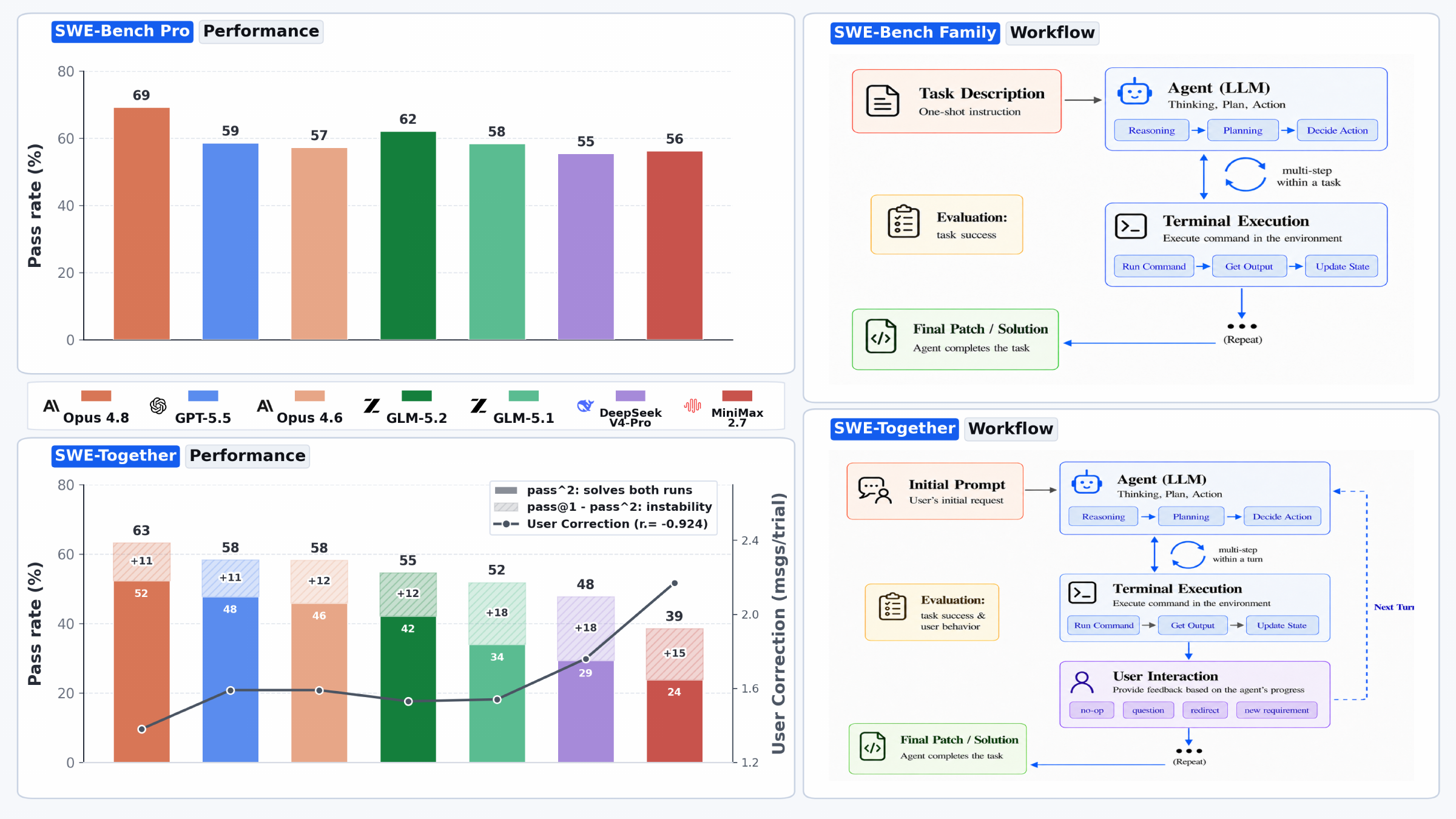}
\captionof{figure}{
    \ours reframes coding-agent evaluation from one-shot SWE-Bench-style tasks into replayable, multi-turn sessions. \textbf{Top:} current mainstream coding benchmarks; \textbf{Bottom:} \ours. \textbf{Left:} pass rates with average user-correction turns overlaid; \textbf{Right:} workflow comparison between existing benchmarks and \ours.
}
\label{fig:teaser}
\par
\endgroup
\vspace{1.0\baselineskip}

\section{Introduction}
\label{sec:intro}
Coding agents are increasingly used as software engineering assistants, and benchmarks for these agents have become prominent measures of capability in frontier model releases. Prominent examples include the SWE-Bench family and Terminal-Bench~\citep{jimenez2024swebench,openai2024sweverified,scale2025swebenchpro,merrill2026terminal}. More broadly, many widely used coding benchmarks follow a static protocol: they present the full task description at the start and evaluate the agent’s submitted code by running executable tests~\citep{chen2021evaluating,austin2021program,jimenez2024swebench,zhuo2025bigcodebench,datacurve2026deepswe}. This protocol has driven substantial progress in the development of coding agents. However, as mainstream coding benchmarks approach saturation and become less able to distinguish among frontier models, there is a growing need for evaluations that better reflect the experience of human users working with coding agents in real workflows.

The mismatch between benchmarks and practice is twofold: task design and evaluation.
First, most benchmarks cast tasks as fixed, single-turn instructions. Real-world coding assistance, by contrast, is inherently interactive: users often reveal intent incrementally, clarify incomplete requests, refine requirements, and correct prior outputs across turns~\citep{zhong2025codechat,zhang2025decodingcoding}. 
Existing benchmarks therefore miss key conditions of real use, where task-relevant information may be distributed across multiple turns and initial requests may be far less complete than benchmark task descriptions~\citep{laban2025lostmultiturn,garg2025savingswebench}.
Second, evaluation primarily measures final-task success, but rarely capture how effectively agents incorporate evolving instructions and how much user effort is required. Two agents may receive the same final-task score while imposing very different burdens: one may succeed from a coarse initial request, while another may require a detailed specification, repeated reminders, and extensive corrective feedback. Evaluating coding agents therefore requires moving beyond final-task correctness to account for interaction quality and the human effort needed for success.

Building such an evaluation entails a fundamental challenge: preserving realistic user interaction while making tasks verifiable. Real-world conversation logs capture how developers use coding agents, but raw sessions are often not benchmark-ready: they may lack a reproducible repository state, an identifiable user goal, or an observable outcome against which success can be assessed. Moreover, the original conversation cannot be replayed verbatim to a new agent: later user turns must be conditioned on the evaluated agent's trajectory, while remaining anchored to the original user's intent; otherwise, the interaction may drift, making final outcomes incomparable. Addressing this challenge therefore requires both verifiable task construction and controlled interaction replay.

To address this challenge, we introduce \ours, a benchmark for evaluating coding agents through multi-turn sessions reconstructed from real user--agent conversations. \ours converts selected sessions into sandboxed tasks by retaining sessions with recoverable repository commits, clear user intents, and concrete outcomes such as submitted code changes. Each task includes the restored repository, the user's initial request as the first-turn instruction, and task-specific artifacts derived from the recorded session, including decomposed user intents and trigger conditions for each feedback turn. During evaluation, an anchored, state-conditional LLM user simulator releases feedback only when its triggering conditions arise in the evaluated agent's trajectory, preserving the original user's objectives and intervention order while adapting feedback timing to each agent. This design helps attribute outcome differences to the agents rather than to simulator variation. Final repository states are scored using task-specific rubrics derived from repository inspection and original-session evidence. Beyond final correctness, \ours reports User Correction, which quantifies the amount of corrective steering provided by the simulated user, and Intent Coverage, which measures whether the simulator consistently communicates the underlying user intents across agent runs. Our contributions are summarized as follows:
\begin{itemize}[topsep=0pt,leftmargin=*,noitemsep]
    \item We introduce \ours, a 109-task benchmark reconstructed from real multi-turn user interactive coding-agent sessions, together with a pipeline that converts recorded sessions into verifiable tasks.
    \item We develop an anchored, state-conditional LLM user simulator that preserves the original user's intent and intervention order while adapting feedback to each evaluated agent's evolving trajectory.
    \item We design a joint evaluation protocol that scores final repository correctness against frozen, implementation-agnostic rubrics and characterizes interaction trajectories through User Correction and Intent Coverage.
\end{itemize}
\section{\ours}
\label{sec:method}
\ours transforms recorded multi-turn coding-agent sessions into reproducible interactive software-engineering tasks.
Our methodology has three components.
First, the \textit{task construction pipeline} filters and normalizes raw sessions, screens whether their coding objectives can be reproduced locally, and converts viable sessions into sandboxed repository-level tasks with pinned environments, executable checks, and task-specific user-simulation prompts.
Second, the \textit{user simulator} replays the original user's intent in a trajectory-conditioned manner, intervening only when conditions derived from the recorded session are satisfied.
Third, the \textit{evaluation framework} separately measures the correctness of the agent's final repository state and the user-simulator behavior during replay.
Correctness is assessed by an agentic rubric judge using repository inspection and executable evidence, while user-simulator behavior is characterized through \textsc{User Correction} and \textsc{Intent Coverage}.
Together, these components enable controlled evaluation of both an agent's ability to complete coding tasks with evolving instructions and the corrective steering it elicits from the simulator.

\subsection{Session-to-Task Construction}
\label{sec:session_collection_pipeline}
We construct executable tasks from raw multi-turn coding sessions drawn from upstream Hugging Face datasets. Four upstream sources contribute to the evaluated suite: DataClaw~\citep{dataclaw2026}, Pi-staging~\citep{pistaging}, Hyperswitch~\citep{hyperswitch}, and SWE-chat~\citep{swechat}, summarized in \tabref{tab:upstream_sources}.

\begin{figure*}[h]
\centering
\includegraphics[width=0.85\linewidth]{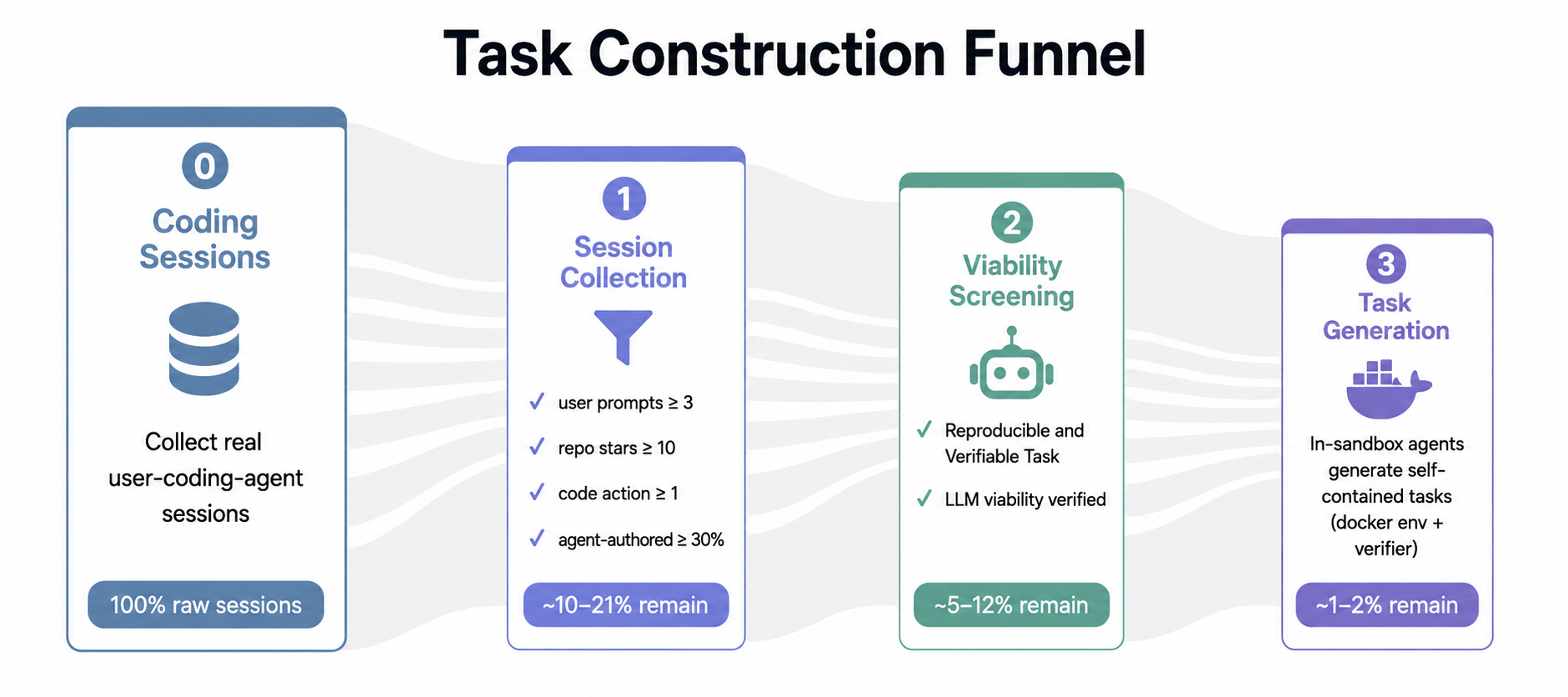}
\caption{
    Overview of the session-to-task construction pipeline.
    The first stage is fully deterministic, the second stage performs viability screening, and the third stage generates reproducible tasks in sandboxes.
}
\label{fig:session_collection_pipeline}
\end{figure*}

\begin{table}[h]
\centering
\caption{
    Upstream sources used to construct \ours, all drawn from real-world user sessions with coding agents.
    ``Final tasks'' denotes sessions that passed eligibility filtering and viability screening, and were successfully converted into executable tasks for the evaluated 109-task suite.
}
\vspace{-0.1in}
\label{tab:upstream_sources}
\begin{tabularx}{\linewidth}{@{}l X r r@{}}
\toprule
Source & Dataset description & Raw sessions & Final tasks \\
\midrule
DataClaw~\citep{dataclaw2026}
  & 32 community-contributed datasets
  & 2{,}228 & 29 \\
Pi-staging~\citep{pistaging}
  & 29 datasets from the Pi staging pipeline
  & 2{,}397 & 23 \\
Hyperswitch~\citep{hyperswitch}
  & Traces from a production payments codebase
  & 784 & 9 \\
SWE-chat~\citep{swechat}
  & Sessions spanning multiple agent harnesses
  & 5{,}851 & 48 \\
\midrule
\textbf{Total} & & \textbf{11{,}260} & \textbf{109} \\
\bottomrule
\end{tabularx}
\end{table}

The final suite is a deliberately high-precision subset of the raw sessions: \(109/11{,}260\) sessions pass the filters and are converted into executable tasks, a conversion rate of \(0.97\%\). Early filters favor public, sufficiently mature GitHub repositories with multi-turn user interaction and concrete code-changing work; later filters require recoverable changes and outcomes that can be evaluated locally.

The construction pipeline has three stages.
First, a deterministic rule-based collector filters raw upstream sessions and emits a normalized record for each candidate.
Second, an LLM judge determines whether the coding work can be reconstructed as a reproducible and verifiable task.
Third, a sandbox orchestrator runs task construction to produce a complete task directory.
The overview is shown in \figref{fig:session_collection_pipeline}.

\subsubsection{Step 1: Deterministic Eligibility Filtering}
\label{subsec:step1_collect}
The first stage constructs an initial pool of candidate sessions using deterministic filtering. 
Given upstream coding-agent sessions, the collector removes traces that lack enough interaction, code modification, or repository context to support replay, and emits one normalized record for each remaining candidate. 
These criteria are deliberately rule-based and require no LLM calls.

A candidate must contain multiple genuine user messages, include concrete agent actions or code edits, and provide enough repository signal to identify the working repository.
The interaction and edit criteria ensure that each trace contains both multi-turn user feedback and concrete code-changing work, rather than only discussion or read-only exploration.
Repository filters favor public, sufficiently mature projects so that downstream sandbox construction is feasible and less dependent on private or unstable codebases.
We additionally filter out sessions in which the final change was primarily authored by the human user.
This preserves trajectories where the coding agent performed substantive implementation work.
%


\begin{figure}[t]
\centering
\includegraphics[width=0.7\linewidth]{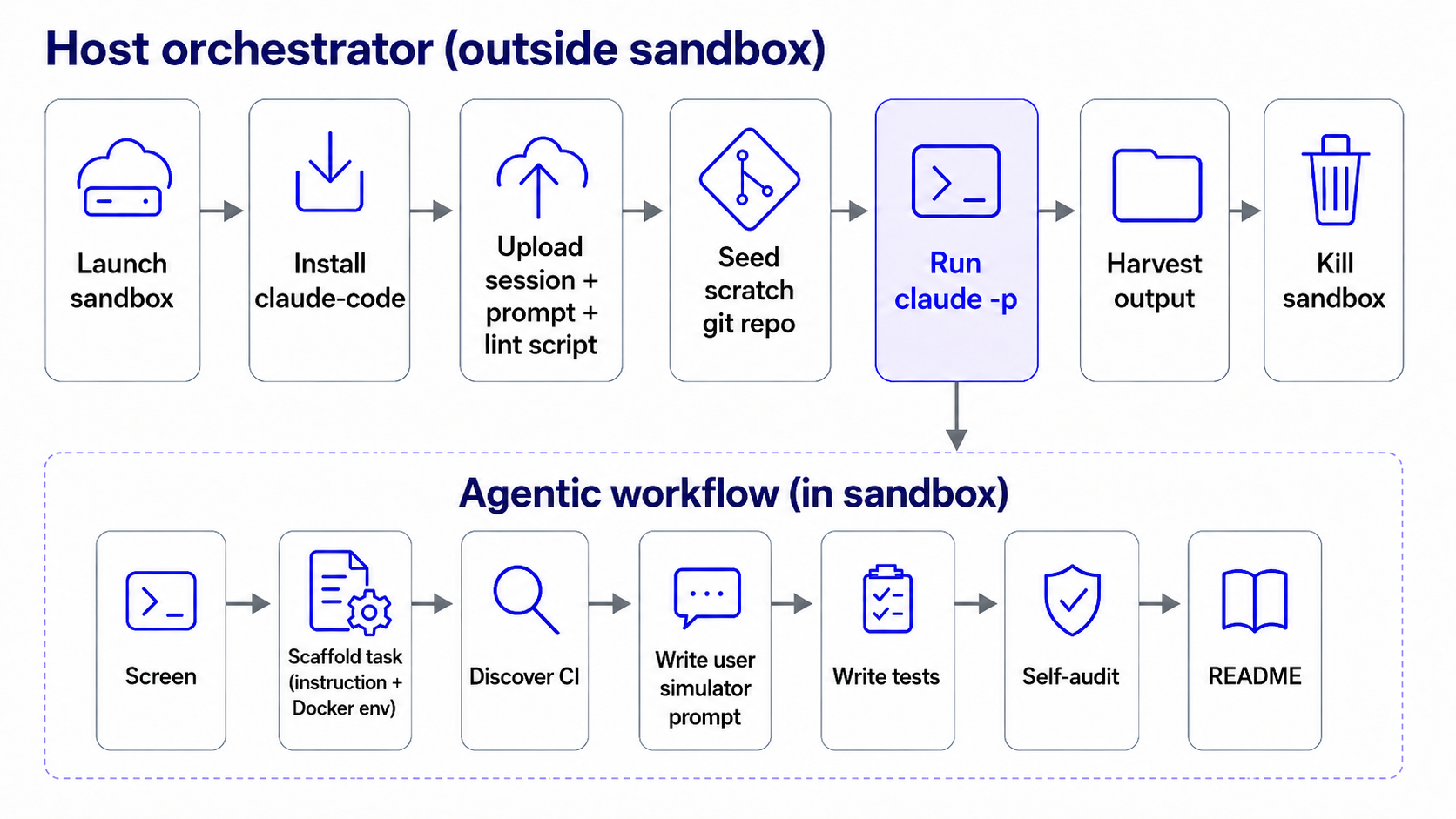}
\caption{
    Task-construction workflow.
    The host orchestrator creates an isolated sandbox, supplies the normalized session and authoring prompt, and exports the resulting task package. 
    Inside the sandbox, the task-generation agent screens the candidate, identifies repository verification commands, writes tests, constructs the user-simulation prompt, and audits the generated task.
}
\label{fig:step3_task_gen}
\end{figure}

\subsubsection{Step 2: Viability Screening}
\label{subsec:step2_screen}
The second stage screens whether the substantive coding work in a retained session can be reconstructed as a self-contained, locally executable task.
The screener receives a compact session summary: repository metadata, message/tool/edit counts, a tool-use distribution, selected user messages, edited file paths, and truncated shell commands. It returns the session's primary deliverable and whether that deliverable is reproducible in the local benchmark environment.

We reject sessions whose primary deliverable depends on external state, such as pull-request management, issue triage, deployment operations, private credentials, or live-service state. Sessions with incidental external actions, such as a final push or pull-request creation step, remain viable when the code edits are the core outcome.
The viability screen does not evaluate correctness. Retained sessions are later reconstructed as original reference patches and evaluated through deterministic verifiers and final repository-state scoring.

\subsubsection{Step 3: Task Construction}
\label{subsec:step3_task_gen}
The third stage converts each viable session into an executable benchmark package.
For each candidate, a host orchestrator launches an isolated sandbox, provides the normalized session and prompt 
and harvests the generated task directory. Inside the sandbox, a task-generation agent performs a stricter repository-grounded screen, clones the target repository at a pinned commit, identifies local setup and test commands, and writes the task artifacts. This separation prevents task construction from depending on host-machine state such as cached credentials, local paths, installed toolchains, or an already-applied fix.

The resulting package contains the original session record, the initial user instruction, a pinned execution environment, deterministic verifier artifacts, and a task-specific user-simulation prompt. \figref{fig:step3_task_gen} summarizes the workflow.

\subsection{User Simulator}
\label{sec:user_agent}
Real software-engineering tasks often evolve through user interventions: clarification, correction, new requirements, or requests to inspect external artifacts. We model this by replaying each reconstructed task as a multi-turn interaction between a coding agent and a user simulator. After each completed agent turn, the replay procedure summarizes the live trajectory and consults the simulator once. The simulator then makes one structured decision: send a user-facing message or remain silent.

The simulator action space contains \textsc{no-op}, \textsc{question}, \textsc{redirect}, \textsc{new-requirement}, and \textsc{check-external}. The default action is \textsc{no-op}, which keeps the simulator silent and lets the agent continue without consuming one of the original follow-up messages. The other actions correspond to common user interventions: asking for clarification, redirecting an unproductive trajectory, introducing a follow-up requirement, or asking the agent to inspect an external artifact.

\begin{figure*}[h]
\centering
\includegraphics[width=0.9\linewidth]{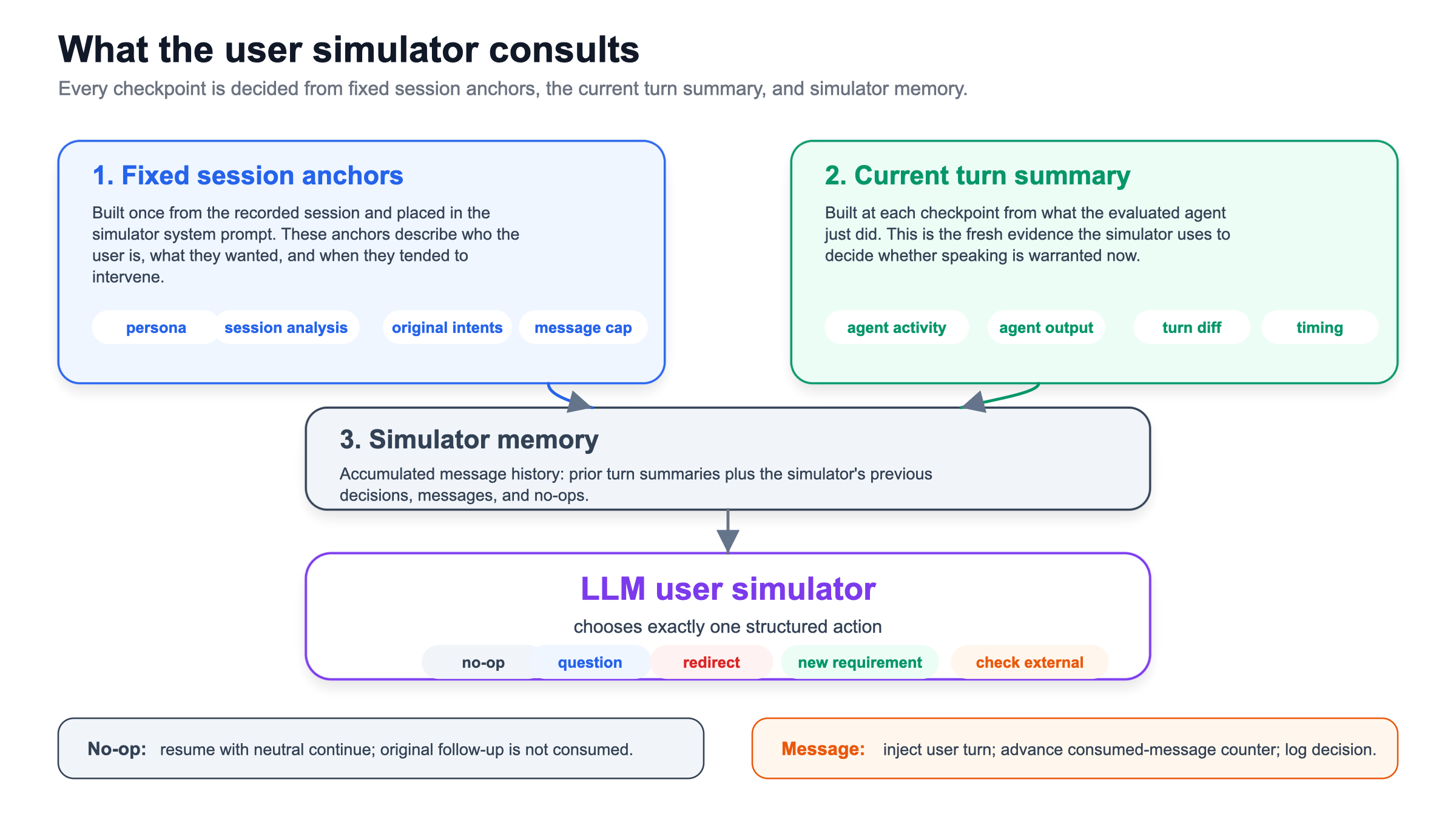}
\caption{
    Context consulted by the user simulator.
    Each replay checkpoint combines fixed session anchors, a summary of the evaluated agent's latest turn, and simulator memory from earlier checkpoints.
    The simulator emits one structured action: message-bearing actions are injected as user turns, while \textsc{no-op} lets the agent continue.
}
\label{fig:user_sim_policy}
\end{figure*}

The simulator follows two principles: interventions are trajectory-conditioned rather than scheduled, and anchored to the original session rather than generic. It conditions on recent agent activity, the agent's latest response, elapsed time, observed repository changes, and its own previous decisions, which helps avoid repeated, premature, or irrelevant messages. At the same time, each task-specific simulator is conditioned on a session analysis reconstructed from the original user--agent interaction~\citep{wu2026humanlm}. This analysis specifies the user's objective, constraints, and intervention conditions grounded in the original follow-up messages. This avoids two failure modes: fixed replay can be mistimed when the evaluated agent follows a different path, while generic simulation can drift away from the original task. At evaluation time, these anchors define a state-conditioned decision policy: the simulator speaks when the live trajectory warrants feedback and otherwise returns \textsc{no-op}.

\subsection{Evaluation Method}
\label{sec:eval_method}
We evaluate each replay along two dimensions: \emph{task correctness} and \emph{user-simulator behavior}. Task correctness measures whether the agent's final repository state satisfies the coding request, including requirements introduced during the interaction. User-simulator behavior measures the feedback needed to produce that final state.

\begin{figure}[h]
\centering
\includegraphics[width=\linewidth]{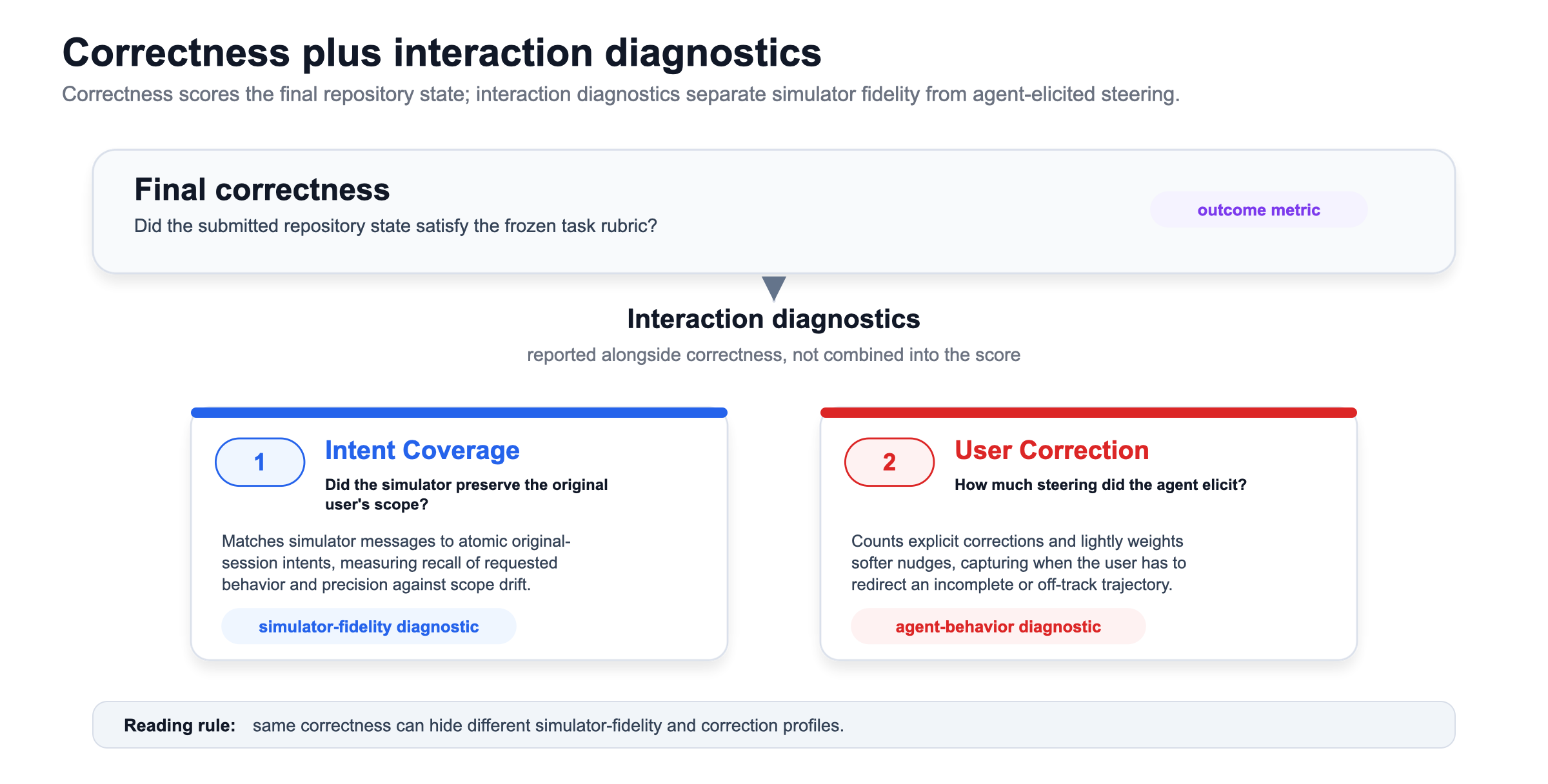}
\caption{
    Correctness plus interaction diagnostics.
    Final correctness scores the submitted repository state, while the interaction diagnostics characterize the replay that produced that state.
    Intent Coverage measures simulator fidelity to the original user's scope; User Correction measures agent-elicited steering.
}
\label{fig:user_metrics}
\end{figure}

\subsubsection{Task Correctness}
\label{sec:task_correctness}
For task correctness, we score behavioral completeness rather than similarity to the original reference patch. This distinction matters because different agents may satisfy the same request with different helper functions, control flow, or integration points.

The executable checks produced during task construction are useful but not sufficient for final scoring. Fixed checks can be misaligned with session intent in both directions: narrow checks may enforce incidental implementation details, while broad checks may require behavior that was never requested. This failure mode is not unique to our setting: OpenAI stopped reporting SWE-bench Verified scores after finding that many remaining tasks used tests that rejected functionally correct solutions or required underspecified behavior~\citep{openai2026swebench}, and DeepSWE reports similar verifier-reliability failures in SWE-bench Pro, including false negatives for behaviorally valid patches~\citep{datacurve2026deepswe}. Execution-only scoring can also miss requirements introduced later in the interaction, fail to exercise behavior that depends on realistic repository context, or reject valid alternative implementations. Purely static review has the opposite weakness: relevant evidence may appear in call sites, configuration, generated behavior, or targeted execution results rather than in a local code fragment alone.

Our evaluator therefore combines deterministic verifiers with an agentic rubric judge. The deterministic verifier provides executable evidence for each task. Separately, the rubric judge scores task completeness in two phases: Phase~1 runs once per task to derive a weighted task rubric, and Phase~2 applies that same rubric to each candidate repository state. We separate the two phases so that the rubric is fixed offline, before and independently of any candidate, preventing the scoring criteria from being tailored to or biased by the particular solution under evaluation and thereby preserving cross-agent comparability. The Phase~1 rubric may consult the original reference patch to identify the behavior of the recorded solution, but the resulting goals are behavioral and are reused unchanged across agents.

For each goal, Phase~2 returns a binary \texttt{met} decision with supporting evidence. The final task-correctness score is derived mechanically from those decisions:
\[
\mathrm{score}
=
\operatorname{round}\!\left(
\sum_g w_g\,\mathbb{I}[g\ \mathrm{met}],
2
\right),
\]
where \(g\) ranges over the task-rubric goals and the weights \(w_g\) are normalized to sum to one. Weighting provides partial credit, while reusing the same rubric across agents ensures that all attempts on the same task are evaluated against identical criteria. A host-side validator checks goal coverage, weight normalization, and consistency between the reported decisions and the resulting score.


\subsubsection{User Simulation Behavior}
Correctness alone does not capture the interaction cost of reaching a solution. Two agents may end with similar repository states while requiring very different amounts of simulator feedback. We therefore report user-simulator behavior separately from task correctness.

We use two diagnostics. Intent Coverage audits simulator fidelity 
: whether replayed simulator messages preserve the original user's intents and remain within scope. User Correction measures the corrective steering elicited by the evaluated agent, counting explicit corrections and lower-weight nudges. This separation keeps simulator fidelity distinct from the agent-facing interaction signal used in the main results.

\textbf{Intent Coverage.}
Intent Coverage measures how faithfully the simulator preserves the original user intents of the original session. We compute it in two passes. First, once per task, we decompose the original session trajectories into a set of atomic original-session intents, each representing a distinct request made by the original user. Second, for each replayed trial, we match the simulator's messages against these intents and record both how well each intent is covered and whether each simulator message remains within the original scope. 

From this matching, we derive weighted intent recall \(I_{\mathrm{recall}}\), which measures how completely the simulator re-expresses the original requests, and scope precision \(I_{\mathrm{precision}}\), which measures how consistently its guidance remains within the original user's scope. We combine them as follows.
\[
\mathrm{IntentCoverage}
=
\operatorname{round}\!\left(
0.70\,I_{\mathrm{recall}}
+
0.30\,I_{\mathrm{precision}},
2
\right).
\]
We weight recall more heavily because omitting an original intent can directly
change the task presented to the evaluated agent. Measuring Intent Coverage
separately helps distinguish agent capability from simulator-induced variation
in how faithfully the original session is reconstructed. The metric therefore
serves as a diagnostic of simulator fidelity and cross-agent comparability:
scores should remain broadly stable across model cohorts, while a substantially
lower score may indicate that the simulator omitted requirements, drifted beyond
the original scope, or struggled to convey the remaining intents because the
agent's responses diverted the interaction from the original trajectory.

\textbf{User Correction.}
We define ``User Correction'' to test the hypothesis that a stronger model requires less intervention to reach the same level of performance. To identify corrective interventions, we apply a multi-label tagger to every simulator message. Multi-labeling is necessary because a message may perform several communicative acts simultaneously, such as introducing a new request while correcting an earlier mistake.

The taxonomy separates three layers of user behavior. The corrective layer contains \texttt{correction}, which explicitly asserts that the agent's work is incorrect, incomplete, or off track, and \texttt{nudge}, which only implies doubt or encourages reconsideration without asserting a defect. The non-corrective ask layer contains \texttt{request}, \texttt{question}, and \texttt{verification}, representing new requirements, genuine information-seeking questions, and neutral checks, respectively. Finally, \texttt{workflow}, \texttt{approval}, and \texttt{context} capture mechanical instructions, confirmations, and background information. These latter tags do not contribute to User Correction.

For each trial, we compute
\[
\mathrm{UserCorrection}
=
N_{\mathrm{correction}}
+
0.2\,N_{\mathrm{nudge}}.
\]
Explicit corrections receive full weight because they directly assert that the agent failed or misunderstood the request. Nudges receive a smaller weight because they capture softer corrective pressure, such as skeptical questions or diagnostic information supplied as a hint. User Correction is first averaged across replicates within each task and then across tasks, ensuring that each task contributes equally to the model-level result. Variation in User Correction across coding agents is primarily agent-dependent and serves as a companion measure to task correctness. Under our hypothesis that stronger models require less corrective intervention to reach comparable performance, User Correction should be negatively correlated with model capability.

\section{Experiments and Results}
\label{sec:experiments}

\subsection{Main Result}
\label{sec:main_result}

\begin{comment}
\begin{figure}[t]
  \centering
  \includegraphics[width=\linewidth]{assets/pass1_ssr_full109_userinput.png}
  \caption{Cumulative pass@1 (left) and stable solve rate (right) as a function
  of the User Input (specification) budget $k$, for the seven opencode cohorts on
  the $109$-task set. The dashed line is the original reference-patch accuracy;
  each curve's endpoint is the corresponding headline value in
  Table~\ref{tab:headline}.}
  \label{fig:pass1-ssr}
\end{figure}
\end{comment}

\textbf{Experimental setup.} We evaluate seven frontier models on the full SWE-Together benchmark of \(109\) tasks using a common agent harness, \texttt{opencode}, with \(k=2\) replicates per task. Each final patch receives a score in \([0,1]\) from the agentic judge against the task's frozen rubric from Section~\ref{sec:task_correctness}. We use this judge score, rather than raw test execution, as the primary correctness signal.

\textbf{Evaluation metrics.}
Let \(j_{t,r}\in[0,1]\) denote the judge score for replicate \(r\) of task \(t\), and let
\[
s_{t,r}=\mathbb{I}[j_{t,r}\ge\tau],\qquad
\bar{j}_t=\frac{1}{k}\sum_{r=1}^{k}j_{t,r},\qquad
\bar{s}_t=\frac{1}{k}\sum_{r=1}^{k}s_{t,r},
\]
We report four equally task-weighted correctness metrics:
\[
\begin{aligned}
\mathrm{pass@1}
&=\frac{1}{N}\sum_{t=1}^{N}\bar{s}_t,
&
\mathrm{SSR}
&=\frac{1}{N}\sum_{t=1}^{N}
  \mathbb{I}[\bar{j}_t\ge\tau],
\\
\mathrm{pass}^{2}
&=\frac{1}{N}\sum_{t=1}^{N}
  \prod_{r=1}^{2}s_{t,r},
&
\mathrm{MeanJudge}
&=\frac{1}{N}\sum_{t=1}^{N}\bar{j}_t.
\end{aligned}
\]
Here, \(N=109\). The three threshold-based metrics apply the same success threshold, \(\tau=0.85\), but aggregate the \(k=2\) runs differently. \(\mathrm{pass@1}\) is the marginal per-run success rate and estimates the probability that a single fresh run solves the task. The stable solve rate (SSR) first averages the continuous judge scores within each task and then applies the threshold, measuring whether the model is reliable on average while tolerating an occasional weak or near-miss run. In contrast, \(\mathrm{pass}^{2}\) measures joint success and credits a task only when both replicates exceed the threshold, providing the strictest measure of consistency and penalizing run-to-run variance most strongly. Consequently, \(\mathrm{pass}^{2} \leq \mathrm{pass@1}\). \(\mathrm{MeanJudge}\) complements these binary metrics by reporting the average continuous judge score without thresholding.

We additionally report User Correction as the interaction diagnostic most directly tied to agent behavior. It measures the corrective steering elicited by an agent and is first averaged over replicates within each task and then across tasks. We report Intent Coverage separately as a simulator-fidelity diagnostic rather than as a model-ranking metric. We also report output-plus-reasoning tokens and wall-clock time per task.

\begin{table}[t]
\centering
\caption{Results on the full \(109\)-task SWE-Together benchmark using the
\texttt{opencode} harness and \(k=2\) replicates. Models are ranked by mean judge score.
Oracle denotes the reference-patch baseline. U-Corr measures corrective
steering (\(\downarrow\) is better), Bold values indicate the best evaluated
agent on the correctness, U-Corr, and efficiency metrics.}
\label{tab:headline}
\small
\setlength{\tabcolsep}{4pt}
\resizebox{\linewidth}{!}{%
\begin{tabular}{rlcccccccc}
\toprule
Rank & Model & pass@1\(\uparrow\) & SSR\(\uparrow\) & pass$^{2}$\(\uparrow\) & Mean judge\(\uparrow\)
& U-Corr\(\downarrow\) &  Tok./task & Min./task \\
\midrule
\(\star\) & Reference
& \(\sim78\%\) & \(\sim78\%\) & \(\sim78\%\) & \(0.90\)
& --- & --- & --- & --- \\
\midrule
1 & Claude Opus 4.8
& \(\mathbf{63\%}\) & \(\mathbf{59\%}\) & \(\mathbf{52\%}\)
& \(\mathbf{0.801}\) & \(\mathbf{1.38}\) &  \(74.0\mathrm{k}\) & \(23.3\) \\
2 & GPT-5.5
& \(58\%\) & \(55\%\) & \(48\%\)
& \(0.763\) & \(1.59\) &  \(\mathbf{29.9\mathrm{k}}\) & \(\mathbf{10.7}\) \\
3 & Claude Opus 4.6
& \(58\%\) & \(58\%\) & \(46\%\)
& \(0.755\) & \(1.59\) & \(42.0\mathrm{k}\) & \(23.2\) \\
4 & GLM-5.2
& \(55\%\) & \(48\%\) & \(42\%\)
& \(0.735\) & \(1.53\) &  \(41.7\mathrm{k}\) & \(24.5\) \\
5 & GLM-5.1
& \(52\%\) & \(49\%\) & \(35\%\)
& \(0.729\) & \(1.54\) &  \(41.6\mathrm{k}\) & \(38.8\) \\
6 & DeepSeek-V4-Pro
& \(48\%\) & \(38\%\) & \(29\%\)
& \(0.679\) & \(1.76\) &  \(49.8\mathrm{k}\) & \(21.0\) \\
7 & MiniMax-2.7
& \(40\%\) & \(34\%\) & \(26\%\)
& \(0.630\) & \(2.17\) & \(43.4\mathrm{k}\) & \(36.2\) \\
\bottomrule
\end{tabular}%
}
\end{table}

\textbf{Overall performance.}
Table~\ref{tab:headline} shows a broadly consistent ordering across the four correctness metrics. Claude Opus 4.8 achieves the strongest overall performance, leading in pass@1 (\(63\%\)), SSR (\(59\%\)), \(\mathrm{pass}^{2}\) (\(52\%\)), and mean judge score (\(0.801\)). It also requires the least corrective steering, with a mean User Correction of \(1.38\). Nevertheless, its pass@1 remains approximately \(15\) percentage points below the original reference-patch accuracy, indicating remaining headroom.

GPT-5.5 ranks second by mean judge score (\(0.763\)), with pass@1 of \(58\%\), SSR of \(55\%\), and \(\mathrm{pass}^{2}\) of \(48\%\); it roughly ties Claude Opus 4.6, with higher \(\mathrm{pass}^{2}\) but lower SSR.
Claude Opus 4.6 follows closely in third (mean judge \(0.755\)). GLM-5.2 and GLM-5.1 form the next tier: their SSR values are nearly identical (\(48\%\) and \(49\%\)), but GLM-5.2 achieves higher pass@1 (\(55\%\) versus \(52\%\)) and a substantially higher \(\mathrm{pass}^{2}\) (\(42\%\) versus \(35\%\)), indicating greater performance stability across replicates. DeepSeek-V4-Pro and MiniMax-2.7 rank last, with MiniMax-2.7 obtaining the lowest values on all four correctness metrics and requiring the most corrective steering. 


\begin{figure}[h]
  \centering
  \includegraphics[width=0.9\linewidth]{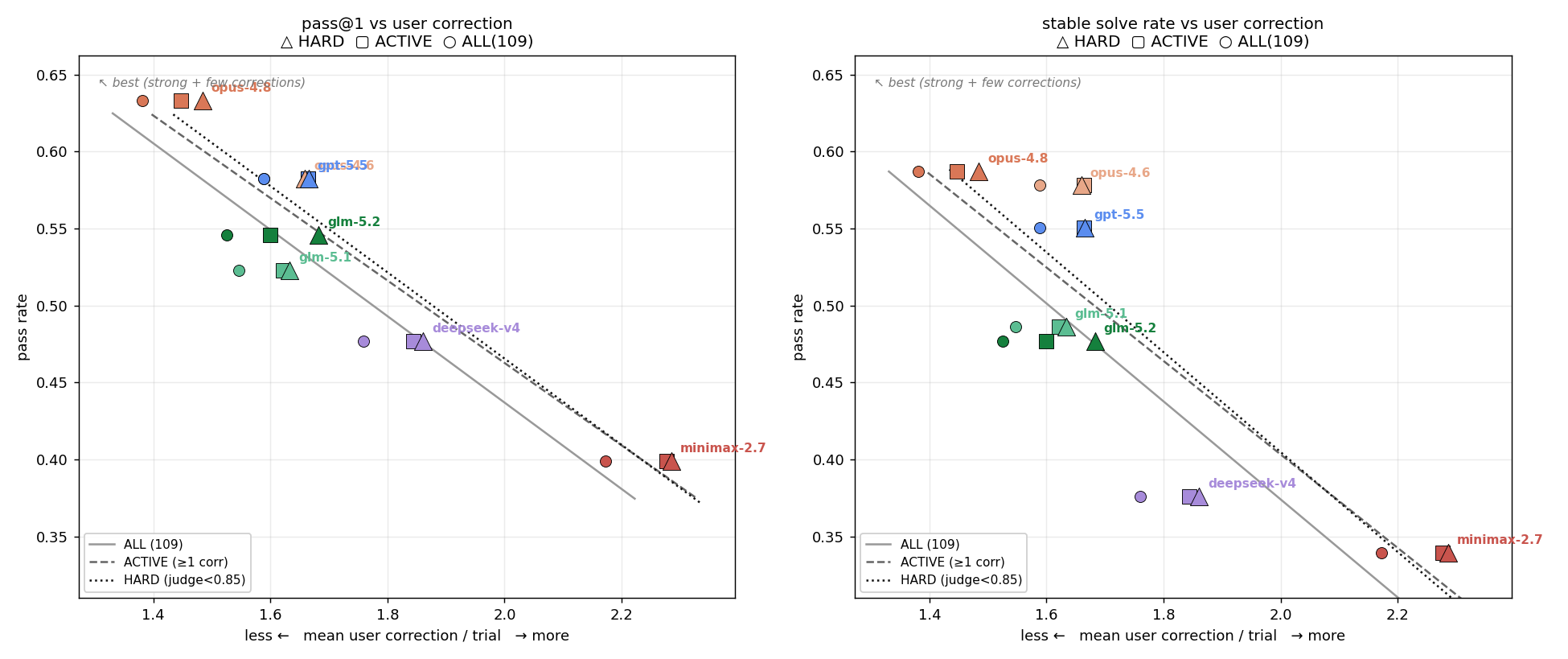}
  \caption{Capability (pass@1 left, stable solve rate right) versus User
  Correction per trial across the seven models, over three task subsets (all
  $109$; the active subset receiving $\ge 1$ correction; the hard subset with
  mean judge $<0.85$). Capability and correction are strongly inversely related
  (Pearson $-0.92$ and $-0.84$ for pass@1 and stable solve rate,
  respectively): stronger models need less corrective pushback.}
  \label{fig:correction}
\end{figure}

\begin{figure}[t]
  \centering
  \includegraphics[width=0.8\linewidth]{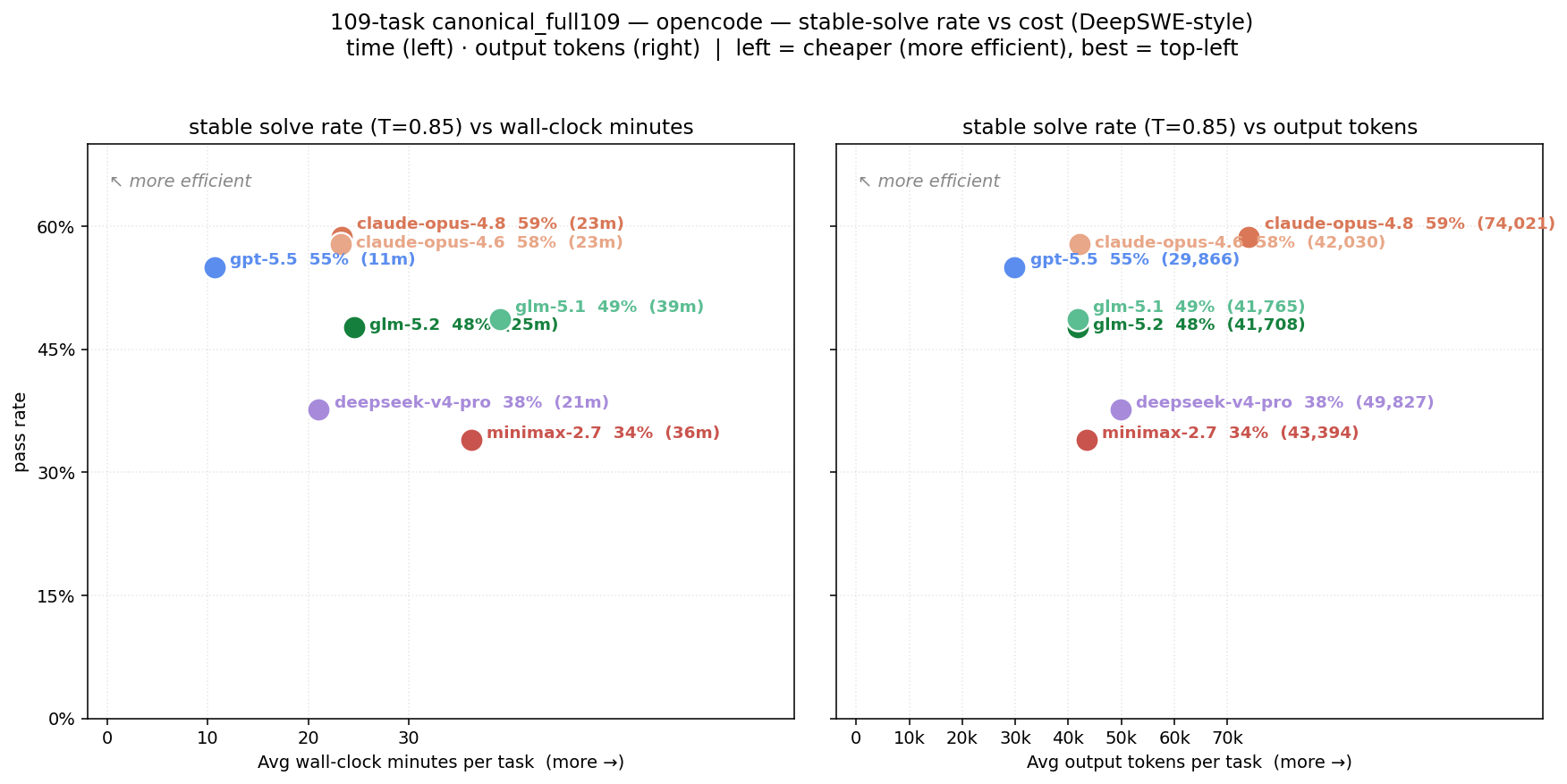}
  \caption{Stable solve rate versus cost for the seven opencode cohorts: mean
  per-task wall-clock minutes (left) and output$+$reasoning tokens (right).
  Up-and-left is better (higher capability at lower cost).}
  \label{fig:efficiency}
\end{figure}

\textbf{Stronger models need less steering.}
User Correction is the one interaction signal that tracks capability: across the
seven models it is strongly \emph{inversely} correlated with pass@1 (Pearson
$-0.92$), stable solve rate ($-0.84$), and mean judge score ($-0.93$). As
Figure~\ref{fig:correction} shows,
opus-4.8 reaches the top of the board with the fewest corrective messages per
trial ($1.38$), whereas minimax-2.7---the weakest cohort---requires the most
($2.17$); the relationship holds on the harder task subsets as well. This
operationalises the intuition that a stronger agent needs less human pushback to
reach the same outcome.

\textbf{Efficiency.}
Figure~\ref{fig:efficiency} plots stable solve rate against per-task wall-clock
time and output tokens. The two cost axes are only weakly coupled to capability
and to each other: gpt-5.5 is the most efficient cohort on both ($29.9$k output tokens
and $10.7$ min per task) while also ranking second on capability, whereas opus-4.8
reaches the top of the board at the highest token cost ($74.0$k) but moderate
wall-clock ($23.3$ min). 
Regarding wall-clock latency, GLM-5.1 and MiniMax are the slowest cohorts, at 38.8 and 36.2 minutes, respectively. However, these latency differences may be affected by the serving location or infrastructure of the inference endpoint.

\textbf{Why the reference scores below 100\%.}
The reference baseline attains a mean judge score of \(0.90\) and a pass rate of
\(\approx\!78\%\) (\(73/93\)) at the \(\tau=0.85\) threshold, evaluated on the
\(93\) of \(109\) tasks that have an extractable reference patch, the remaining
\(16\) have no canonical code diff.
Of the \(93\), \(57\) (\(61\%\)) score a perfect \(1.0\) and \(73\) (\(78\%\)) pass; only
\(20\) fall below pass threshold, and their shortfall has three identifiable sources
rather than task unresolvability. First, roughly 35\% of the unsatisfied
goals are \emph{process} requirements that the rubric inherits from the recorded
session---diagnosing the root cause before editing, answering a follow-up
question, or explaining the change to the user---which a final patch cannot
express and which therefore penalize \emph{every} submission, reference and agent
alike. Second, a few reference patches incompletely capture work that originally
spanned multiple commits or sessions, an extraction-noise artifact of the data
pipeline. Third, a small residual reflects genuinely
imperfect human solutions---e.g., a session that fixes several code paths but
leaves the headline bug, which the judge correctly identifies. Since the same frozen rubric is applied to the reference patch and to every evaluated agent, these shared factors largely affect all submissions alike, and the reference row is best read as a like-for-like reference point under our scoring criteria rather than a strict ceiling on resolvability.

\subsection{User Simulator Consistency Analysis}
\label{sec:user-sim}
Intent Coverage remains broadly stable across coding agents. Six of the seven model cohorts obtain overall scores between $0.70$ and $0.72$, with recall ranging from $0.72$ to $0.74$ and precision from $0.66$ to $0.72$. This narrow variation indicates that the simulator generally preserves the original users' requests and remains within their intended scope, despite differences in the behavior of the underlying coding agents. The resulting evaluations are therefore broadly comparable across model cohorts and are unlikely to be driven by systematic variation in simulator fidelity.

GPT-5.5 sits marginally below this band, with an overall score of $0.68$ (recall $0.71$, precision $0.65$). The gap is small and consistent with its strong, efficient task performance: a more capable agent more often resolves or reshapes the interaction on its own, which can leave the simulator fewer natural openings to re-express every original follow-up, rather than indicating reduced simulator fidelity for this cohort. All seven cohorts exhibit consistent user-simulator behavior within a tight range, supporting cross-agent comparability.

\subsection{User Simulator Quality Study}
\label{sec:user-study}

\textbf{Protocol.}
We evaluate whether human annotators can distinguish simulated users from real users. Through a web interface, four annotators make forced two-alternative-choice judgments over paired trajectories, selecting the trajectory they believe was produced by the real user. We use the 52 tasks shared by DeepSeek-V4-Pro, MiniMax-2.7, and Claude Opus 4.6, yielding 156 trajectory pairs and 312 judgments.

We report the \emph{Turing pass rate}, defined as the fraction of judgments in which the simulated trajectory is selected as real. A pass rate of $50\%$ corresponds to chance-level discrimination.

\textbf{Findings.}
Across all trajectories, the simulator achieves a Turing pass rate of $46\%$ (95\% CI $[40.5,\,51.6]\%$). Because the confidence interval includes $50\%$, annotators exhibit no statistically reliable ability to distinguish simulated users from real users. These results indicate that, under our evaluation protocol, simulated and real user trajectories are generally indistinguishable to human evaluators.

\section{Related Work}\label{sec:related}
We distinguish two forms of multi-turn evaluation for coding agents.
Agent-environment multi-turn refers to episodes in which an agent iteratively inspects files, runs commands, edits code, and invokes tests while solving a fixed user request.
%
Interactive replay refers to episodes in which the user-facing instruction evolves through feedback,corrections, clarifications, or new requirements. 
This distinction organizes prior work into three groups: 
\emph{(i)} software-engineering benchmarks with realistic agent--environment interaction but fixed user requests; 
\emph{(ii)} interactive coding benchmarks with simulated users; 
and \emph{(iii)} real coding-session datasets without task-grounded replay.
%
As summarized in \tabref{tab:benchmark_positioning}, \ours combines repository-level agent-environment interaction, interactive user-correction replay, and provenance from real recorded user-agent coding sessions.
\begin{table}[h]
\centering
\caption{
    Positioning of \ours against representative coding-agent benchmarks.
    \cmark indicates yes, \xmark indicates no, \pmark indicates partial support, and \mmark indicates mixed or heterogeneous coverage.
    %
    Agent-env. multi-turn means the agent can iteratively interact with tools, files, terminal, or tests. 
    Interactive replay means evaluation includes sequential user-facing feedback, correction, or clarification turns.
}
\label{tab:benchmark_positioning}
\renewcommand{\arraystretch}{1.12}
\setlength{\tabcolsep}{12pt}
\resizebox{\textwidth}{!}{%
\begin{tabular}{lccccc}
\toprule
\textbf{Benchmark} &
\textbf{\shortstack{Repo-\\level}} &
\textbf{\shortstack{Agent-env.\\multi-turn}} &
\textbf{\shortstack{Interactive\\replay}} &
\textbf{\shortstack{Real task\\source}} &
\textbf{\shortstack{Real user\\session}} \\
\midrule

SWE-bench family~\citep{jimenez2024swebench,pan2024swegym,liang2026swenext}
& \cmark & \cmark & \xmark & \cmark & \xmark \\

Terminal-Bench~\citep{merrill2026terminal}
& \xmark & \cmark & \xmark & \xmark & \xmark \\

MINT / ConvCodeWorld~\citep{wang2024mint,han2025convcodeworld}
& \xmark & \pmark & \cmark & \xmark & \xmark \\

CodeAssistBench~\citep{kim2025codeassistbench}
& \cmark & \cmark & \cmark & \cmark & \xmark \\

RECODE-H / FronTalk~\citep{miao2025recodeh,wu2026frontalk}
& \pmark & \pmark & \cmark & \pmark & \xmark \\

BigCodeArena / CodeChat~\citep{zhuo2025bigcodearena,zhong2025codechat}
& \mmark & \mmark & \pmark & \cmark & \cmark \\

SWE-chat~\citep{swechat}
& \cmark & \cmark & \xmark & \cmark & \cmark \\

\midrule
\textbf{\ours (ours)}
& \cmark & \cmark & \cmark & \cmark & \cmark \\
\bottomrule
\end{tabular}}
\end{table}

\textbf{Agent-environment multi-turn benchmarks.}
SWE-bench~\citep{jimenez2024swebench} and its extensions~\citep{yang2025swesmith,zhang2025swebenchlive,liang2026swenext,pan2024swegym} are the canonical repository-level software-engineering benchmarks. 
They ground tasks in real codebases and issues, and agents may take many environment actions before producing a final patch. 
Terminal-Bench~\citep{merrill2026terminal} similarly evaluates agents in interactive terminal environments. 
These benchmarks are multi-turn in the agent--environment sense, but the user request is fixed: the benchmark does not evaluate whether an agent can recover from user corrections or adapt to requirements revealed after intermediate attempts.
%
\ours keeps the repository-level, tool-using setting, but adds an interactive user-correction loop.

\textbf{Interactive coding benchmarks with simulated users.}
A complementary line of work introduces simulated user feedback into coding evaluation. 
TiCoder~\citep{lahiri2022ticoder} studies iterative code generation with simulated user queries on HumanEval~\citep{chen2021evaluating} and MBPP~\citep{austin2021program}.
MINT~\citep{wang2024mint} and ConvCodeWorld~\citep{han2025convcodeworld} extend LLM-driven feedback to standard code-generation tasks.
And~\citet{pan2025whenbenchmarkstalk} converts static benchmarks into interactive evaluations by revealing hidden information through a simulated user.
More recent work moves interactive evaluation closer to software engineering: CodeAssistBench (CAB)~\citep{kim2025codeassistbench} uses GitHub-issue tasks and a satisfaction-driven simulated user.
SWEET-RL/ColBench~\citep{zhou2025sweetrl}, RECODE-H~\cite{miao2025recodeh}, and FronTalk~\cite{wu2026frontalk} evaluate collaborative refinement in backend, research-code, and front-end settings, and CollabLLM~\citep{wu2025collabllm} trains assistants to act as active collaborators by optimizing multi-turn-aware rewards estimated through a simulated user.
These benchmarks show that user feedback is an important evaluation axis, but their interaction loops are generally synthesized from static tasks, issue artifacts, curated feedback policies, or benchmark-generated scenarios~\citep{xu2025funreasonmt,zhou2026sandmle,chen2026dreamgym}.
In contrast, \ours derives tasks from recorded real human--agent coding sessions and grounds the simulated correction loop in the corresponding original session.
This design choice is supported by \citet{suh2026simulators}, who quantify the utility of user simulators and find that simulators grounded in real human behavior yield substantially better downstream collaborative assistants than role-playing prompts.

\textbf{Real coding-session data.}
Recent datasets demonstrate that real user--agent coding interactions can be collected at scale. 
SWE-chat~\citep{swechat} characterizes real-world coding-agent sessions, while BigCodeArena~\cite{zhuo2025bigcodearena} and CodeChat~\cite{zhong2025codechat} collect code-centric conversations for preference modeling, analysis, and assistant evaluation. 
Related benchmarks such as Saving SWE-Bench~\cite{garg2025savingswebench} and EDIT-Bench~\cite{chi2025editbench} further emphasize realistic user phrasing and in-the-wild edit instructions. 
However, these works are primarily descriptive, preference-oriented, or single-request evaluation sets: they do not pair each real session with a replayable repository state, deterministic verifier, and live user-correction loop for evaluating new agents.
%
\section{Limitations and Conclusion}
\label{sec:limitations}
\textbf{Limitation.}
The user simulator cannot interrupt the coding agent during its turns, cannot directly edit files, and relies solely on textual trajectories and tool outputs rather than visual information from the interface. The current design works best when user goals and constraints are clearly defined and focuses on tasks with measurable outcomes, such as submitted patches. Consequently, it provides limited coverage of ambiguous, open-ended tasks and qualitative user behaviors that are difficult to quantify.

\textbf{Conclusion.} 
This work introduces SWE-Together, a benchmark that transforms real multi-turn coding sessions into reproducible software-engineering tasks and evaluates both final task correctness and the user guidance required during interaction. Across 109 tasks and seven frontier models, the results show substantial differences in capability and reliability. User Correction is strongly negatively correlated with performance, supporting the hypothesis that more capable coding agents require less user intervention. In addition, the simulator maintains broadly consistent intent coverage across models and produces trajectories that human annotators cannot reliably distinguish from real-user interactions. Together, these findings demonstrate the importance of evaluating coding agents not only by whether they complete a task, but also by how much corrective steering they elicit. By incorporating realistic multi-turn interactions and user-centered diagnostics, SWE-Together aims to evaluate coding agents in a way that more closely reflects the actual user experience.

\section*{Acknowledgments}
We thank Zhiqing Sun and Rui Hou for insightful discussions and feedback.

\ours is built entirely on coding-agent sessions that were collected, curated, and
openly released by the opensource community. We are deeply grateful to the trajectory
providers whose donated data made this benchmark possible: the DataClaw
community~\citep{dataclaw2026}, the contributors behind the Pi staging
pipeline~\citep{pistaging}, the maintainers of the Hyperswitch trace
collection~\citep{hyperswitch}, and the SWE-chat work~\citep{swechat}. 

\bibliographystyle{assets/plainnat}
\bibliography{reference}

@article{chen2021evaluating,
  title={Evaluating large language models trained on code},
  author={Chen, Mark and Tworek, Jerry and Jun, Heewoo and Yuan, Qiming and Pinto, Henrique Ponde De Oliveira and Kaplan, Jared and Edwards, Harri and Burda, Yuri and Joseph, Nicholas and Brockman, Greg and others},
  journal={arXiv preprint arXiv:2107.03374},
  year={2021}
}

@article{austin2021program,
  title={Program synthesis with large language models},
  author={Austin, Jacob and Odena, Augustus and Nye, Maxwell and Bosma, Maarten and Michalewski, Henryk and Dohan, David and Jiang, Ellen and Cai, Carrie and Terry, Michael and Le, Quoc and others},
  journal={arXiv preprint arXiv:2108.07732},
  year={2021}
}

@inproceedings{wang2024mint,
  author    = {Xingyao Wang and Zihan Wang and Jiateng Liu and Yangyi Chen and Lifan Yuan and Hao Peng and Heng Ji},
  title     = {{MINT}: Evaluating {LLMs} in Multi-turn Interaction with Tools and Language Feedback},
  booktitle = {International Conference on Learning Representations (ICLR)},
  year      = {2024}
}

@misc{xu2025funreasonmt,
  author = {Zengzhuang Xu and Bingguang Hao and Zechuan Wang and Yuntao Wen and Xinyi Xu and Yang Liu and Long Chen and Dong Wang and Maolin Wang and Tong Zhao and Yicheng Chen and Cunyin Peng and Jinjie Gu and Leilei Gan and Xiangyu Zhao and Chenyi Zhuang and Shi Gu},
  title  = {{FunReason-MT} Technical Report: Advanced Data Synthesis Solution for Real-world Multi-Turn Tool-use},
  year   = {2025},
  eprint = {2510.24645},
  archivePrefix = {arXiv},
  primaryClass = {cs.AI}
}

@misc{chen2026dreamgym,
  author = {Zhaorun Chen and Zhuokai Zhao and Kai Zhang and Bo Liu and Qi Qi and Yifan Wu and Tarun Kalluri and Sara Cao and Yuanhao Xiong and Haibo Tong and Huaxiu Yao and Hengduo Li and Jiacheng Zhu and Xian Li and Dawn Song and Bo Li and Jason Weston and Dat Huynh},
  title  = {{DreamGym}: Scaling Agent Learning via Experience Synthesis},
  year   = {2026},
  eprint = {2511.03773},
  archivePrefix = {arXiv},
  note   = {ICLR 2026}
}

@misc{zhou2026sandmle,
  author = {Yuhang Zhou and Lizhu Zhang and Yifan Wu and Jiayi Liu and Xiangjun Fan and Zhuokai Zhao and Hong Yan},
  title  = {Synthetic Sandbox for Training Machine Learning Engineering Agents},
  year   = {2026},
  eprint = {2604.04872},
  archivePrefix = {arXiv},
  primaryClass = {cs.LG}
}

@inproceedings{han2025convcodeworld,
  author    = {Hojae Han and Seung-won Hwang and Rajhans Samdani and Yuxiong He},
  title     = {{ConvCodeWorld}: Benchmarking Conversational Code Generation in Reproducible Feedback Environments},
  booktitle = {International Conference on Learning Representations (ICLR)},
  year      = {2025}
}

@misc{kim2025codeassistbench,
  author = {Myeongsoo Kim and Shweta Garg and Baishakhi Ray and Varun Kumar and Anoop Deoras},
  title  = {{CodeAssistBench (CAB)}: Dataset and Benchmarking for Multi-turn Chat-Based Code Assistance},
  year   = {2025},
  eprint = {2507.10646},
  archivePrefix = {arXiv}
}

@misc{laban2025lostmultiturn,
  author = {Philippe Laban and Hiroaki Hayashi and Yingbo Zhou and Jennifer Neville},
  title  = {{LLMs} Get Lost In Multi-Turn Conversation},
  year   = {2025},
  eprint = {2505.06120},
  archivePrefix = {arXiv}
}

@misc{wu2026humanlm,
  author = {Shirley Wu and Evelyn Choi and Arpandeep Khatua and Zhanghan Wang and Joy He-Yueya and Tharindu Cyril Weerasooriya and Wei Wei and Diyi Yang and Jure Leskovec and James Zou},
  title  = {{HumanLM}: Simulating Users with State Alignment Beats Response Imitation},
  year   = {2026},
  eprint = {2603.03303},
  archivePrefix = {arXiv}
}

@inproceedings{wu2025collabllm,
  author    = {Shirley Wu and Michel Galley and Baolin Peng and Hao Cheng and Gavin Li and Yao Dou and Weixin Cai and James Zou and Jure Leskovec and Jianfeng Gao},
  title     = {{CollabLLM}: From Passive Responders to Active Collaborators},
  booktitle = {International Conference on Machine Learning (ICML)},
  year      = {2025}
}

@misc{zhou2025sweetrl,
  author = {Yifei Zhou and Song Jiang and Yuandong Tian and Jason Weston and Sergey Levine and Sainbayar Sukhbaatar and Xian Li},
  title  = {{SWEET-RL}: Training Multi-Turn {LLM} Agents on Collaborative Reasoning Tasks},
  year   = {2025},
  eprint = {2503.15478},
  archivePrefix = {arXiv}
}

@misc{scale2025swebenchpro,
  author = {{Scale AI}},
  title  = {{SWE-Bench Pro}: A New Frontier for {SWE-Agents}},
  year   = {2025},
  howpublished = {\url{https://scale.com/research/swe_bench_pro}}
}

@misc{zhong2025codechat,
  author = {Suzhen Zhong and Ying Zou and Bram Adams},
  title  = {Developer-{LLM} Conversations: An Empirical Study of Interactions and Generated Code Quality},
  year   = {2025},
  eprint = {2509.10402},
  archivePrefix = {arXiv}
}

@misc{zhang2025decodingcoding,
  author = {Binquan Zhang and Li Zhang and Haoyuan Zhang and Fang Liu and Song Wang and Bo Shen and An Fu and Lin Shi},
  title  = {Decoding Human-{LLM} Collaboration in Coding: An Empirical Study of Multi-Turn Conversations in the Wild},
  year   = {2025},
  eprint = {2512.10493},
  archivePrefix = {arXiv}
}

@misc{garg2025savingswebench,
  author = {Spandan Garg and Benjamin Steenhoek and Yufan Huang},
  title  = {Saving {SWE-Bench}: A Benchmark Mutation Approach for Realistic Agent Evaluation},
  year   = {2025},
  eprint = {2510.08996},
  archivePrefix = {arXiv}
}

@inproceedings{jimenez2024swebench,
  author    = {Carlos E. Jimenez and John Yang and Alexander Wettig and Shunyu Yao and Kexin Pei and Ofir Press and Karthik Narasimhan},
  title     = {{SWE-bench}: Can Language Models Resolve Real-World {GitHub} Issues?},
  booktitle = {International Conference on Learning Representations (ICLR)},
  year      = {2024}
}

@misc{openai2024sweverified,
  author       = {{OpenAI}},
  title        = {Introducing {SWE-bench Verified}},
  year         = {2024},
  howpublished = {\url{https://openai.com/index/introducing-swe-bench-verified/}}
}

@misc{pan2024swegym,
  author = {Jiayi Pan and Xingyao Wang and Graham Neubig and Navdeep Jaitly and Heng Ji and Alane Suhr and Yizhe Zhang},
  title  = {Training Software Engineering Agents and Verifiers with {SWE-Gym}},
  year   = {2024},
  eprint = {2412.21139},
  archivePrefix = {arXiv}
}

@inproceedings{yang2025swesmith,
  author    = {John Yang and Kilian Lieret and Carlos E. Jimenez and Alexander Wettig and Kabir Khandpur and Yanzhe Zhang and Binyuan Hui and Ofir Press and Ludwig Schmidt and Diyi Yang},
  title     = {{SWE-smith}: Scaling Data for Software Engineering Agents},
  booktitle = {Advances in Neural Information Processing Systems (NeurIPS) Datasets and Benchmarks Track},
  year      = {2025}
}

@inproceedings{zhang2025swebenchlive,
  author    = {Linghao Zhang and Shilin He and Chaoyun Zhang and Yu Kang and Bowen Li and Chengxing Xie and Jianfeng Wang and Maoquan Wang and Yufan Huang and Shengyu Fu and Elsie Nallipogu and Qingwei Lin and Yingnong Dang and Saravan Rajmohan and Yudong Zhang},
  title     = {{SWE-bench} Goes Live!},
  booktitle = {Advances in Neural Information Processing Systems (NeurIPS) Datasets and Benchmarks Track},
  year      = {2025}
}

@inproceedings{zhuo2025bigcodebench,
  author    = {Terry Yue Zhuo and Minh Chien Vu and Jenny Chim and Han Hu and Wenhao Yu and Ratnadira Widyasari and Imam Nur Bani Yusuf and Haolan Zhan and Junda He and Indraneil Paul},
  title     = {{BigCodeBench}: Benchmarking Code Generation with Diverse Function Calls and Complex Instructions},
  booktitle = {International Conference on Learning Representations (ICLR)},
  year      = {2025}
}

@misc{dataclaw2026,
  author       = {{DataClaw Community}},
  title        = {{DataClaw}: A Donated Corpus of Real Coding-Agent Sessions},
  year         = {2026},
  howpublished = {\url{https://huggingface.co/datasets/alexshengzhili/dataclaw-harbor-candidates}}
}

@software{pistaging,
  author       = {Zechner, Mario},
  title        = {pi-share-hf: Collect, review, and upload redacted pi session files to a Hugging Face dataset},
  year         = {2026},
  publisher    = {GitHub},
  howpublished = {\url{https://github.com/badlogic/pi-share-hf}},
}

@misc{hyperswitch,
  author       = {Archit11},
  title        = {claude\_traces\_hs},
  year         = {2026},
  publisher    = {Hugging Face},
  howpublished = {\url{https://huggingface.co/datasets/archit11/claude_traces_hs}},
}

@misc{swechat,
      title={SWE-chat: Coding Agent Interactions From Real Users in the Wild}, 
      author={Joachim Baumann and Vishakh Padmakumar and Xiang Li and John Yang and Diyi Yang and Sanmi Koyejo},
      year={2026},
      eprint={2604.20779},
      archivePrefix={arXiv},
      primaryClass={cs.AI},
      url={https://arxiv.org/abs/2604.20779}, 
}

@misc{suh2026simulators,
  author       = {Joseph Suh and Ayush Raj and Minwoo Kang and Serina Chang},
  title        = {Quantifying the Utility of User Simulators for Building Collaborative {LLM} Assistants},
  year         = {2026},
  eprint       = {2605.09808},
  archivePrefix = {arXiv},
  primaryClass = {cs.CL},
  url          = {https://arxiv.org/abs/2605.09808}
}

@misc{lahiri2022ticoder,
  author = {Shuvendu K. Lahiri and Sarah Fakhoury and Aaditya Naik and Georgios Sakkas and Saikat Chakraborty and Madanlal Musuvathi and Piali Choudhury and Curtis von Veh and Jeevana Priya Inala and Chenglong Wang and Jianfeng Gao},
  title  = {Interactive Code Generation via Test-Driven User-Intent Formalization ({TiCoder})},
  year   = {2022},
  eprint = {2208.05950},
  archivePrefix = {arXiv}
}

@misc{pan2025whenbenchmarkstalk,
  author = {Jane Pan and Ryan Shar and Jacob Pfau and Ameet Talwalkar and He He and Valerie Chen},
  title  = {When Benchmarks Talk: Re-Evaluating Code {LLMs} with Interactive Feedback},
  year   = {2025},
  eprint = {2502.18413},
  archivePrefix = {arXiv}
}

@misc{miao2025recodeh,
  author = {Chunyu Miao and Henry Peng Zou and Yangning Li and Yankai Chen and Yibo Wang and Fangxin Wang and Yifan Li and Wooseong Yang and Bowei He and Xinni Zhang and Dianzhi Yu and Hanchen Yang and Hoang H. Nguyen and Yue Zhou and Jie Yang and Jizhou Guo and Wenzhe Fan and Chin-Yuan Yeh and Panpan Meng and Liancheng Fang and Jinhu Qi and Wei-Chieh Huang and Zhengyao Gu and Yuwei Han and Langzhou He and Yuyao Yang and Yinghui Li and Hai-Tao Zheng and Xue Liu and Irwin King and Philip S. Yu},
  title  = {{RECODE-H}: A Benchmark for Research Code Development with Interactive Human Feedback},
  year   = {2025},
  eprint = {2510.06186},
  archivePrefix = {arXiv},
  note   = {Under review at ICLR 2026}
}

@misc{wu2026frontalk,
  author = {Xueqing Wu and Zihan Xue and Da Yin and Shuyan Zhou and Kai-Wei Chang and Nanyun Peng and Yeming Wen},
  title  = {{FronTalk}: Benchmarking Front-End Development as Conversational Code Generation with Multi-Modal Feedback},
  year   = {2026},
  eprint = {2601.04203},
  archivePrefix = {arXiv},
  note   = {Under review at ICLR 2026}
}

@misc{liang2026swenext,
  author = {Jiarong Liang and Zhiheng Lyu and Zijie Liu and Xiangchao Chen and Ping Nie and Kai Zou and Wenhu Chen},
  title  = {{SWE-Next}: Scalable Real-World Software Engineering Tasks for Agents},
  year   = {2026},
  eprint = {2603.20691},
  archivePrefix = {arXiv}
}

@misc{chi2025editbench,
  author = {Wayne Chi and Valerie Chen and Ryan Shar and Aditya Mittal and Jenny Liang and Wei-Lin Chiang and Anastasios Nikolas Angelopoulos and Ion Stoica and Graham Neubig and Ameet Talwalkar and Chris Donahue},
  title  = {{EDIT-Bench}: Evaluating {LLM} Abilities to Perform Real-World Instructed Code Edits},
  year   = {2025},
  eprint = {2511.04486},
  archivePrefix = {arXiv}
}

@misc{zhuo2025bigcodearena,
  author = {Terry Yue Zhuo and Xiaolong Jin and Hange Liu and Juyong Jiang and Tianyang Liu and Chen Gong and Bhupesh Bishnoi and Vaisakhi Mishra and Marek Suppa and Noah Ziems and Saiteja Utpala and Ming Xu and Guangyu Song and Kaixin Li and Yuhan Cao and Bo Liu and Zheng Liu and Sabina Abdurakhmanova and Wenhao Yu and Mengzhao Jia and Jihan Yao and Kenneth Hamilton and Kumar Shridhar and Minh Chien Vu and Dingmin Wang and Jiawei Liu and Zijian Wang and Qian Liu and Binyuan Hui and Meg Risdal and Ahsen Khaliq and Atin Sood and Zhenchang Xing and Wasi Uddin Ahmad and John Grundy and David Lo and Banghua Zhu and Xiaoning Du and Torsten Scholak and Leandro von Werra},
  title  = {{BigCodeArena}: Unveiling More Reliable Human Preferences in Code Generation via Execution},
  year   = {2025},
  eprint = {2510.08697},
  archivePrefix = {arXiv}
}

@article{merrill2026terminal,
  title={Terminal-bench: Benchmarking agents on hard, realistic tasks in command line interfaces},
  author={Merrill, Mike A and Shaw, Alexander G and Carlini, Nicholas and Li, Boxuan and Raj, Harsh and Bercovich, Ivan and Shi, Lin and Shin, Jeong Yeon and Walshe, Thomas and Buchanan, E Kelly and others},
  journal={arXiv preprint arXiv:2601.11868},
  year={2026}
}

@misc{openai2026swebench,
  author       = {{OpenAI}},
  title        = {Why {SWE-bench Verified} No Longer Measures Frontier Coding Capabilities},
  year         = {2026},
  month        = feb,
  day          = {23},
  howpublished = {\url{https://openai.com/index/why-we-no-longer-evaluate-swe-bench-verified/}}
}

@misc{datacurve2026deepswe,
  author       = {Wenqi Huang and Charley Lee and Leonard Tng and Serena Ge},
  title        = {{DeepSWE}: Measuring Frontier Coding Agents on Original, Long-Horizon Engineering Tasks},
  year         = {2026},
  month        = may,
  day          = {26},
  howpublished = {\href{https://deepswe.datacurve.ai/blog/deepswe}{Datacurve blog}},
}


\end{document}